\documentclass{article} 
\usepackage{iclr2027_conference,times}

\usepackage{amsmath,amsfonts,bm}

\def\eqref#1{equation~\ref{#1}}

\def\1{\bm{1}}

\DeclareMathAlphabet{\mathsfit}{\encodingdefault}{\sfdefault}{m}{sl}
\SetMathAlphabet{\mathsfit}{bold}{\encodingdefault}{\sfdefault}{bx}{n}

\usepackage{hyperref}
\usepackage{url}
\usepackage{booktabs}
\usepackage{multirow}
\usepackage{graphicx}
\usepackage{amsmath}
\usepackage{amssymb}
\usepackage{algorithm}
\usepackage{algorithmic}
\usepackage{enumitem}
\usepackage{microtype}
\usepackage{caption}
\usepackage{subcaption}
\usepackage{tcolorbox}

\title{DynSTEER: Dynamic Stage-wise Trajectory Evaluation and Execution-time Review for Agents}

\author{Zhichao Shi$^{1,2,3,4}$, 
Wenjie Zhang$^{5}$, 
Xuhui Jiang$^{3,4}$\thanks{Corresponding author}, 
Xiaojun Wu$^{3,4,6}$, 
Cehao Yang$^{3,4,6}$, \\
\textbf{Chengjin Xu}$^{3,4}$\textbf{,} 
\textbf{Jian Guo}$^4$\textbf{,} \textbf{Yuanzhuo Wang}$^2$\footnotemark[1] \\
$^1$ School of Advanced Interdisciplinary Sciences, UCAS \\
$^2$ State Key Lab of AI Safety, Institute of Computing Technology, CAS \\
$^3$ DataArc Tech Ltd. \\
$^4$ IDEA Research, International Digital Economy Academy \\
$^5$ Jiangnan University \\
$^6$ The Hong Kong University of Science and Technology (Guangzhou)
}

\arxiv

\begin{document}

\maketitle

\begin{abstract}
Large language model agents are increasingly deployed for long-horizon task execution, raising a central granularity question for trajectory evaluation: whole-trajectory verification is too coarse to capture concrete failures and their associated evidence in long trajectories, while atomic-step scoring is too fine-grained, noise-sensitive, and computationally expensive.
This granularity gap makes a single-reference trajectory paradigm inadequate for assessing the rich space of valid agent execution paths and delays timely feedback and early stopping in long-horizon tasks.
To address these issues, we propose DynSTEER, a dynamic stage-wise framework for agent trajectory evaluation.
DynSTEER bridges the granularity gap through stage-wise dynamic evaluation that segments rollouts at key execution nodes and adapts its multi-level review strategy based on stage-level results; it compiles a path-tolerant milestone graph from available task inputs to preserve diverse legal paths without reference leakage; and it supports terminating unrecoverable agent executions to curb resource waste.
Experimental results show that DynSTEER improves evaluation discriminability by over 85\% compared with whole-trajectory evaluation and saves 17.74\% of execution steps. The code is available at 
\url{https://github.com/zhichao-stone/DynSTEER} 
\end{abstract}

\section{Introduction}
\label{sec:introduction}

LLM agents are increasingly undertaking complex, long-horizon tasks across real-world environments~\citep{yao2023react,shinn2023reflexion,jimenez2024swebench,gu2024toolsandbox,zhou2024webarena}.
As task horizons expand, evaluating only final output results becomes insufficient to identify agent deficiencies and diagnose failure modes, therefore bringing trajectory-level evaluation to the forefront of agent research~\citep{qian2024agentprocessbench,shi2025earlyeval}.

However, existing trajectory evaluation paradigms face a central granularity problem, especially for long-horizon tasks.
Whole-trajectory verification is too coarse to focus on concrete failures and their associated evidence in long trajectories, whereas atomic-step scoring is too fine-grained: isolated messages lack task context, so ordinary retries and harmless observations become noise.
This granularity gap further raises two challenges.
Agent executions often admit multiple valid and effective paths. Because of this granularity gap, evaluation typically offers only limited path references and therefore struggles to faithfully reflect the quality of diverse agent executions.
Moreover, full-trajectory review delays diagnosis that could be made earlier, so an execution that has already gone wrong may continue, wasting model resources and triggering additional skill or tool calls.

\begin{figure*}[t]
\centering
\includegraphics[width=0.96\textwidth]{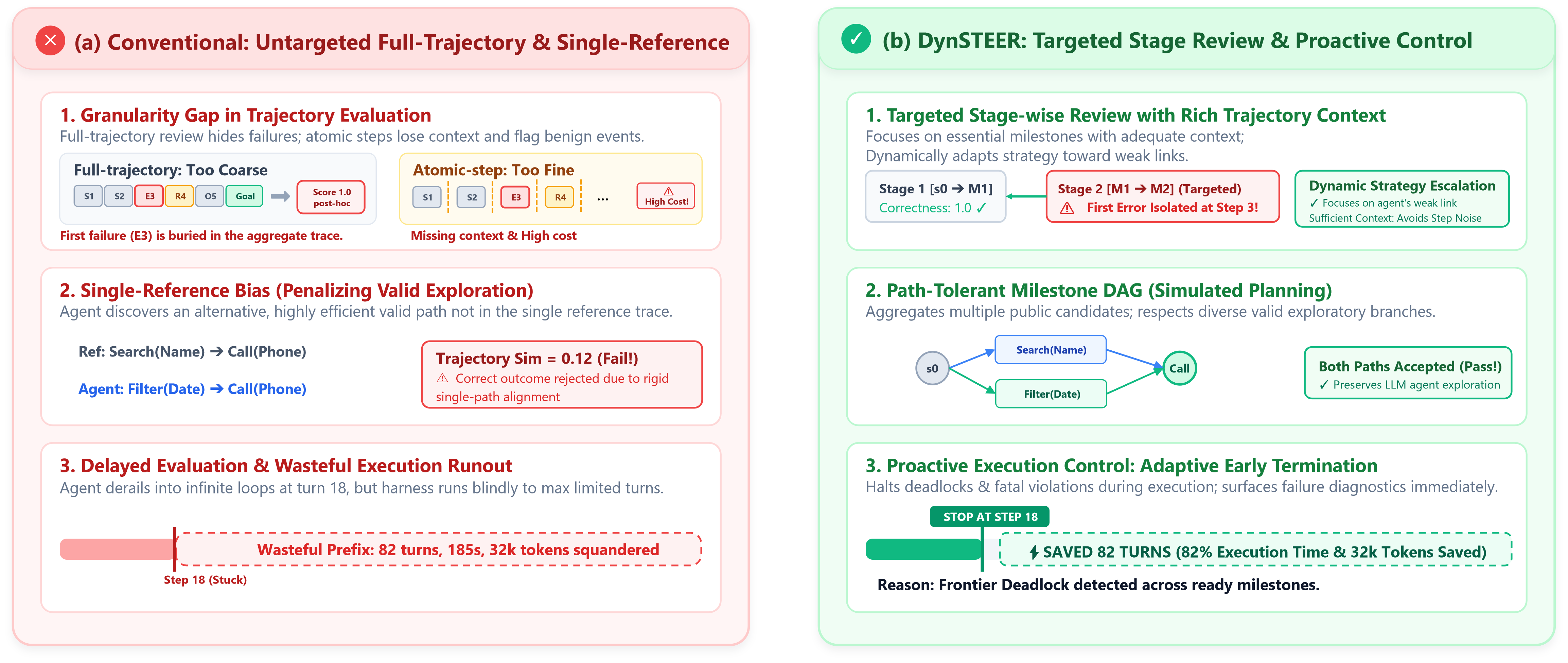}
\vspace{-0.25cm}
\caption{\textbf{Comparison of Agent Evaluation Paradigms.} (a) Conventional paradigms evaluate either coarse outcomes or noisy atomic steps, often against a single reference, and discover waste only after execution. (b) \textsc{DynSTEER} evaluates path-tolerant milestone-bounded stages and uses the resulting stage evidence to target review and stop hopeless prefixes early.}
\label{fig:teaser_comparison}
\end{figure*}

As illustrated in Figure~\ref{fig:teaser_comparison}, we introduce \textbf{DynSTEER}, a \textsc{Dyn}amic \textbf{S}tage-wise \textbf{T}rajectory \textbf{E}valuation and \textbf{E}xecution-time \textbf{R}eview framework.
To address the granularity challenge, DynSTEER performs stage-wise dynamic evaluation: it segments rollouts at necessary execution nodes, preserves task context while isolating localized evidence for error attribution, and adapts its multi-level review strategy based on stage-level results.
To avoid single-reference bias, it compiles a path-tolerant milestone graph from available task inputs via multi-candidate simulated planning and majority consensus, capturing necessary task invariants while accommodating diverse legitimate paths.
Finally, it detects unrecoverable prefixes and terminates them to curb resource waste.

Experiments on ToolSandbox and SWE-bench Pro show that DynSTEER improves discriminability by over 85\% compared with whole-trajectory evaluation and saves 17.74\% and 8.40\% of execution steps.
Comparisons with expert-annotated graphs from ToolSandbox demonstrate the validity of DynSTEER’s compiled milestone graphs.
Extensive ablation studies further confirm the effectiveness of its dynamic adaptation and early-stopping mechanisms in better distinguishing the capabilities of different models as agent backbones.

In summary, our contributions are as follows:
\begin{itemize}[leftmargin=*,itemsep=1.5pt,topsep=1pt]
    \item \textbf{Stage-wise Dynamic Evaluation:} We segment rollouts at necessary execution-step nodes and adapt multi-level review routing from stage-level results, isolating context-rich evidence windows for multidimensional scoring and first-error localization.
    \item \textbf{Path-Tolerant Task Blueprint Compilation:} We compile a milestone DAG from available task inputs via multi-candidate planning and strict-majority consensus, preserving diverse valid paths without reference leakage.
    \item \textbf{Counterfactual Early Termination:} We detect unrecoverable prefixes from stage-level evidence and terminate them, substantially reducing wasted computation.
\end{itemize}

\section{Related Work}
\label{sec:related_work}

\subsection{Agent Trajectory Evaluation}
Unlike outcome benchmarks that inspect only terminal code patches or database records~\citep{jimenez2024swebench,qin2023toolllm,liu2023agentbench,zhou2024webarena}, trajectory evaluation examines intermediate planning decisions, tool invocations, and error recoveries generated during agent execution. Recent benchmarks expand assessment: BFCL~\citep{patil2025bfcl} evaluates multi-step tool calls via AST checks; TRAJECT-Bench~\citep{trajectbench2025} assesses tool selection and dependency order; AgentRewardBench~\citep{agentrewardbench2025} benchmarks trajectory judges; and DeepRed~\citep{deepred2026} and Claw-Eval~\citep{claweval2026} leverage environment snapshots to award partial credit along execution traces.

Despite providing intermediate visibility, existing evaluators face two key limitations. First, they predominantly rely on a single reference path~\citep{qian2024agentprocessbench}. In open environments, tasks frequently admit multiple valid solution routes; single-path matching inherently penalizes legitimate exploratory strategies. Second, they operate post-hoc across the full rollout and lack dynamic intervention to adapt evaluation strategies to an agent's specific weak links.

\subsection{Dynamic Evaluation for Agents}
Static benchmarks suffer from prompt overfitting, data contamination, and leaderboard score saturation. To address these vulnerabilities, researchers have begun exploring dynamic evaluation paradigms. For conversational agents, Agent-Testing Agent~\citep{agenttestingagent2026} adaptively generates subsequent test cases guided by prior judge feedback. In interactive tool environments, ToolSandbox~\citep{gu2024toolsandbox} introduces stateful conversational simulations with pre-defined milestones and safety minefields to evaluate arbitrary rollouts.

However, existing dynamic evaluation approaches primarily concentrate on dynamically selecting or perturbing benchmark test suites~\citep{gu2024toolsandbox,agenttestingagent2026}. During the actual execution of an individual task, trajectory evaluation remains largely passive: evaluators typically check simple milestone reachability at the end of rollouts, lacking a formal mechanism to dynamically steer evaluation focus, update dimension weights, or control execution progression in real time.

\subsection{Efficient Evaluation and Early Stopping Strategies}
As agent rollouts grow longer, evaluation costs have escalated dramatically, costing hundreds of dollars per benchmark pass~\citep{shi2025earlyeval}. Prior efficiency efforts centered on benchmark distillation, subsampling static suites into smaller representative subsets~\citep{aslam2006statistical,shi2025earlyeval}. To improve within-task efficiency, EarlyEval~\citep{shi2025earlyeval} introduces early outcome prediction, training dual LightGBM classifiers on prefix features to predict binary task success/failure and halt execution early.

Nevertheless, benchmark distillation reduces the task count while leaving long, wasteful rollouts untouched, whereas tabular early prediction models~\citep{shi2025earlyeval} act as black boxes. They predict binary outcomes without understanding domain semantics, cannot provide causal failure attribution, and lack awareness of tool dependency constraints or hazardous minefields.

In summary, existing trajectory evaluators either inspect isolated atomic steps that suffer from local noise, or evaluate full traces post-hoc against limited reference paths that penalize valid diversity.
Meanwhile, early stopping methods lack domain semantics and failure attribution.
To bridge this divide, we propose DynSTEER. DynSTEER performs stage-wise dynamic evaluation that adapts review capacity to weak or ambiguous stages. It compiles path-tolerant milestone graphs from available task inputs. Finally, it uses execution-time early stopping to eliminate wasteful computation.

\section{Methodology}
\label{sec:methodology}

\begin{figure*}[t]
\centering
\includegraphics[width=0.96\textwidth]{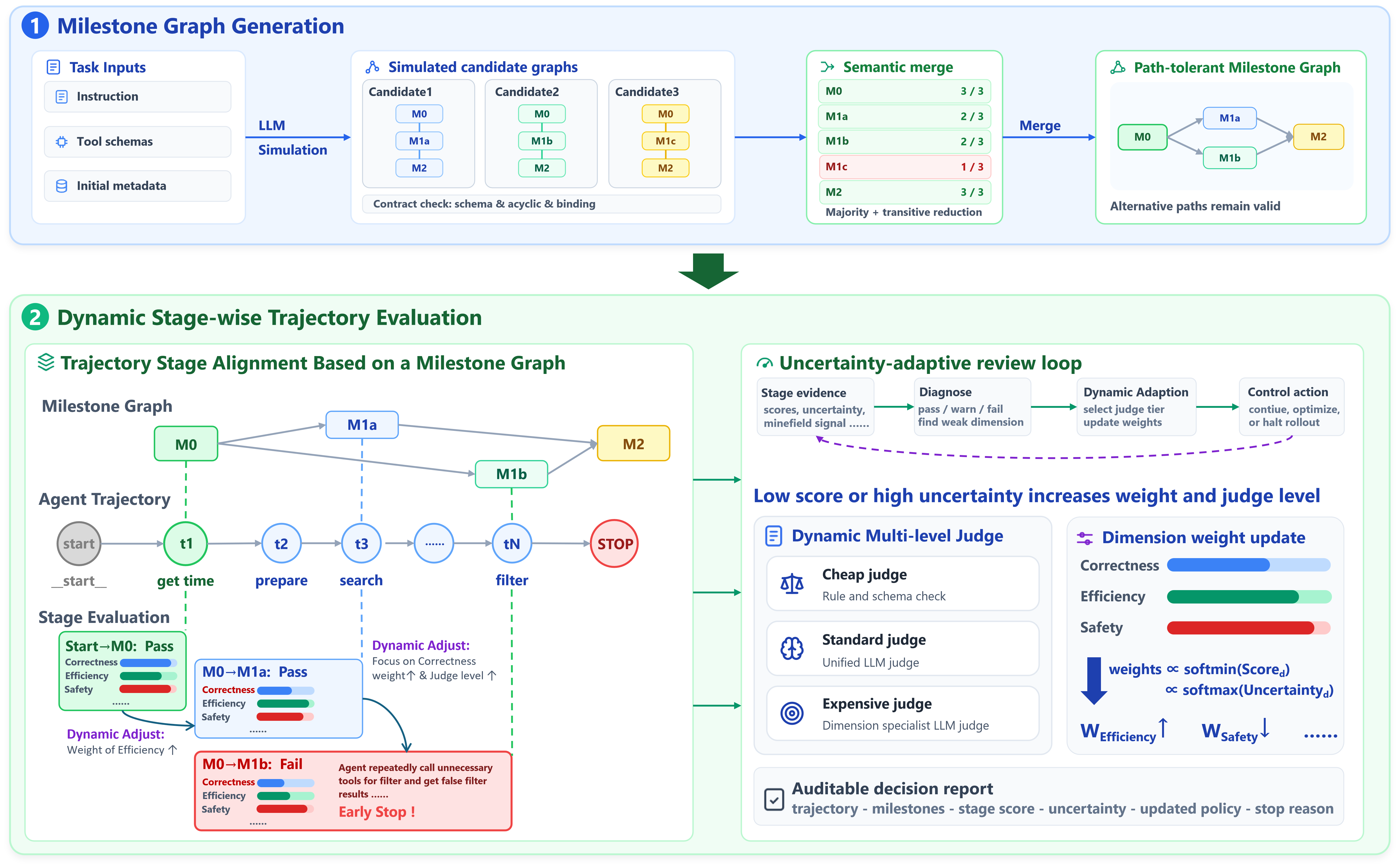}
\vspace{-0.25cm}
\caption{\textbf{Overview of the \textsc{DynSTEER} Workflow.} (1) Pre-execution milestone graph generation synthesizes candidate graphs from task inputs to compile a path-tolerant graph. (2) Dynamic stage-wise evaluation aligns trajectory and milestone graphs to evaluate weak links with trajectory context, adaptively scales judge levels, and records auditable early-stop decisions.}
\label{fig:overall_framework}
\vspace{-0.35cm}
\end{figure*}

DynSTEER establishes a dynamic evaluation framework for agent trajectories, as illustrated in Figure~\ref{fig:overall_framework}. Rather than treating rollouts as opaque black boxes or enforcing rigid single-path matching, DynSTEER coordinates three core components: stage-wise dynamic evaluation to resolve the evidence-granularity problem, which Sections~\ref{sec:stage_alignment} and~\ref{sec:routing_early_stop} present; path-tolerant milestone-graph generation to preserve execution diversity, which Section~\ref{sec:blueprint_compilation} presents; and execution-time early termination to curb resource waste, which Section~\ref{sec:routing_early_stop} presents.

\subsection{Task Inputs and Dual-Clock Observation}
\label{sec:formulation}
Following interactive tool-using agent environments~\citep{gu2024toolsandbox,yang2023intercode,liu2023agentbench}, a task provides instruction $x$, initial state $\mathcal{S}_0$, API catalog $\mathcal{A}$, and safety contracts $\mathcal{C}$. DynSTEER evaluates the task inputs:
\begin{equation}
    V_{\text{task}} = (x, \mathcal{A}, \mathcal{I}_0),
    \label{eq:task_inputs}
\end{equation}
where $\mathcal{I}_0$ is evaluator-visible initial-state metadata; hidden reference paths $\tau^*$ and terminal unit tests are never used for graph construction.

During execution, the trace $\tau = \langle e_1, \dots, e_K \rangle$ is observed on two clocks. 
The raw event clock $k$ indexes every observed event and triggers immediate fatal-action interception, whereas the closed-action clock $t$ advances only after a tool return is observed and the agent yields control. Milestone settlement and scoring use $t$. 
For example, in $\langle \text{tool call}, \text{tool return}, \text{agent yield} \rangle$, the three observations receive raw indices $k=1,2,3$, so a hazardous tool call can be intercepted at $k=1$. 
The action closes at $k=3$, where $t$ advances from $0$ to $1$ and becomes the unit used for milestone settlement. 
This dual-clock design couples safety responsiveness with semantic completeness: the raw event clock exposes fatal minefield operations as soon as the agent attempts them, whereas the closed-action clock prevents milestone settlement from missing information or state changes.

\subsection{Milestone Graph Generation via Simulated Planning}
\label{sec:blueprint_compilation}
To overcome single-reference bias, DynSTEER compiles $V_{\text{task}}$ into a path-tolerant Milestone Graph $G = (M, E, \mathcal{K}, \mathcal{M}_{\text{fatal}})$ prior to execution, where $M$ is the milestone set, $E$ encodes precedence constraints, $\mathcal{K}$ specifies parameter bindings, and $\mathcal{M}_{\text{fatal}}$ defines hazardous operations.

\textbf{Multi-Candidate Synthesis and Contract Validation.} Given $V_{\text{task}}$, a blueprint generator synthesizes $C$ candidate plans $\{G_1, \dots, G_C\}$ via LLM prompting across diverse exploration heuristics, such as direct keyword queries versus hierarchical filtering. Each candidate $G_c = (M_c, E_c)$ undergoes deterministic contract verification:
\begin{equation}
    \text{Valid}(G_c) = \text{SchemaCheck}(M_c, \mathcal{A}) \land \text{Acyclic}(E_c) \land \text{BindingConsistency}(\mathcal{K}_c)
    \label{eq:valid_gc}
\end{equation}
Here, SchemaCheck requires every milestone and minefield to name an available API and to match its argument and return schema; Acyclic rejects circular precedence constraints; BindingConsistency requires dynamic bindings to reference declared upstream outputs with compatible parameter types. A candidate failing any check is discarded. Appendix~\ref{app:contract_validation} details these deterministic filters.

\textbf{Majority Consensus and Transitive Reduction.} To retain essential invariants while preserving exploratory flexibility, a milestone $m$ is retained only if it appears in a majority of valid candidates:
\begin{equation}
    M = \left\{ m \;\middle|\; \sum_{c=1}^{C_{\text{valid}}} \mathbb{I}[m \in M_c] \ge \left\lceil \frac{C_{\text{valid}} + 1}{2} \right\rceil \right\}
    \label{eq:consensus}
\end{equation}
Precedence edges are aggregated analogously. Finally, we compute the transitive reduction $E = \text{TR}(E_{\text{majority}})$~\citep{tarjan1972depth} to eliminate redundant shortcut dependencies, producing a minimal directed acyclic graph that accommodates diverse legitimate execution strategies.

\subsection{Stage-wise Dynamic Trajectory Evaluation}
\label{sec:stage_alignment}
During execution, DynSTEER dynamically aligns closed actions against the milestone graph. Algorithm~\ref{alg:dynsteer} in Appendix~\ref{app:algorithms} presents the complete algorithmic flow.

\textbf{Ready Frontier Matching.} Let $M_{<t}$ denote milestones settled prior to closed action $t$. The active search space is strictly bounded by the \textbf{ready frontier}:
\begin{equation}
    \mathcal{F}_t = \{ m \in M \setminus M_{<t} \mid \text{Pred}(m) \subseteq M_{<t} \}
    \label{eq:frontier}
\end{equation}
An action $a_t$ settles milestone $m^* \in \mathcal{F}_t$ if its tool call and state delta satisfy semantic predicates $\phi_{m^*}(a_t, s_t) = 1$. Restricting alignment to $\mathcal{F}_t$ prevents out-of-order settlement and reward gaming.

\textbf{Dominator-Anchored Stage Isolation.} To construct a semantically complete stage evidence window for $m^*$, DynSTEER computes the immediate dominator $\text{idom}(m^*)$ in $G$~\citep{lengauer1979fast}, defined as the unique node that must be traversed on every directed path from start node $s_0$ to $m^*$. The \textbf{stage evidence window} is bounded by:
\begin{equation}
    \mathcal{W}(m^*) = [t_{\text{idom}(m^*)}, t]
    \label{eq:stage_window}
\end{equation}
where $t_{\text{idom}(m^*)}$ is the execution step at which the dominator was settled. 
Anchoring evidence to $\mathcal{W}(m^*)$ preserves all intermediate reasoning, trial-and-error attempts, and parameter adjustments that lead directly to $m^*$, ensuring that relevant actions are not omitted merely because the agent executes them in parallel branches.
Unlike whole-trajectory evaluation, which obscures mistake locations, and single-step evaluation, which lacks context, dominator-anchored stages provide enough trajectory context to evaluate intermediate quality and pinpoint the exact first-error step.

\textbf{Multidimensional Scoring.} DynSTEER scores the entire evidence window $\mathcal{W}(m^*)$. Each stage uses a task-relevant subset $\mathcal{D}_j \subseteq \mathcal{D}$ of seven dimensions: progress, state consistency, tool quality, efficiency, safety, interaction quality, and recovery. Let $j$ denote the evaluated-stage clock, independent of the closed-action clock $t$; stage $j$ ends at the closed action at which a milestone candidate is evaluated. The composite stage score is:
\begin{equation}
    S_j = \sum_{d \in \mathcal{D}_j} w_d^{(j)} s_d^{(j)}(m^*)
    \label{eq:stage_scoring}
\end{equation}
parameterized by dynamically normalized weights $\mathbf{w}^{(j)}$.
Uncertainty-adaptive multi-level review then treats these stage scores and uncertainty estimates as control signals, allocating stronger review capacity to ambiguous or weak dimensions. 

\subsection{Uncertainty-Adaptive Review and Execution-Time Early Termination}
\label{sec:routing_early_stop}
Uncertainty-adaptive review is the control layer for stage-wise evaluation, adjusting dimension weights and judge levels after each stage. 
Execution-time early termination is a separate downstream safeguard that reuses the same stage evidence to control resource waste.

\textbf{Multi-Level Judge Structure.} Rather than assigning a uniform evaluator across all stages, DynSTEER deploys a three-level judge hierarchy whose prompt templates are detailed in Appendix~\ref{app:prompts}:
\begin{itemize}[leftmargin=*,itemsep=1.5pt,topsep=1pt]
    \item \textbf{Level 1: Cheap Judge} ($J_{\text{cheap})}$: A deterministic structural evaluator inspecting tool status codes, schemas, and evidence rules without LLMs, incurring zero model inference overhead.
    \item \textbf{Level 2: Standard Judge} ($J_{\text{std})}$: A multi-field unified LLM evaluator scoring all target dimensions simultaneously within a single pass, aggregating confidence across multiple samples.
    \item \textbf{Level 3: Expensive Judge} ($J_{\text{exp})}$: A single-dimension specialist LLM evaluator executing dedicated expert rubrics and deep counterfactual analysis across multiple deliberation passes to capture subtle behavioral defects.
\end{itemize}

\textbf{Uncertainty-Adaptive Review Adjustment.} DynSTEER dynamically shifts evaluation weights and escalates judge levels toward an agent's vulnerable links. First, dimension weights are updated via exponential gradient scaling:
\begin{equation}
    w_d^{(j+1)} = \frac{w_d^{(j)} \cdot \exp\left( \alpha (1 - s_d^{(j)}) + \beta u_d^{(j)} \right)}{\sum_{d' \in \mathcal{D}} w_{d'}^{(j)} \cdot \exp\left( \alpha (1 - s_{d'}^{(j)}) + \beta u_{d'}^{(j)} \right)}
    \label{eq:weight_update}
\end{equation}
where $s_d^{(j)} \in [0, 1]$ is the dimension score, $u_d^{(j)} \in [0, 1]$ is dimension uncertainty, $F_j^{\text{fatal}}$ indicates fatal minefield evidence in stage $j$, and $\alpha, \beta$ scale sensitivity to quality deficits and uncertainty. Appendix~\ref{app:weighting_rationale} gives the design rationale and mathematical derivation of this update. Second, dimensions with low scores or high uncertainty receive increased attention in the next-stage evaluation, which is formalized as follows: the baseline judge level is:
\begin{equation}
    \text{Level}_{\text{base}}^{(j+1)} =
    \begin{cases}
    J_{\text{cheap}}, & \text{if } S_j \ge \theta_{\text{pass}} \;\land\; \max_{d} u_d^{(j)} \le \theta_{\text{low\_u}} \;\land\; F_j^{\text{fatal}} = 0 \\
    J_{\text{std}}, & \text{otherwise}
    \end{cases}
    \label{eq:level_base}
\end{equation}
For each individual dimension $d$, its level $\text{Level}_d^{(j+1)}$ adaptively escalates along $J_{\text{cheap}} < J_{\text{std}} < J_{\text{exp}}$:
\begin{equation}
    \text{Level}_d^{(j+1)} =
    \begin{cases}
    J_{\text{exp}}, & \text{if } s_d^{(j)} < \theta_{\text{fail}} \\
    \min\left(J_{\text{exp}},\, \text{Level}_d^{(j)} + 1\right), & \text{if } s_d^{(j)} < \theta_{\text{warn}} \lor u_d^{(j)} \ge \theta_{\text{high\_u}} \\
    \text{Level}_{\text{base}}^{(j+1)}, & \text{otherwise}
    \end{cases}
    \label{eq:level_dim}
\end{equation}
This focuses expensive LLM deliberation selectively on weak or ambiguous execution links while relaxing confident stages to rule-based verification to minimize unnecessary LLM invocations.
Together with stage segmentation and local evidence isolation, this evaluation strategy resolves the evidence-granularity problem. 

\textbf{Execution-Time Early Termination.} To prevent runaway executions on unrecoverable trajectories, DynSTEER monitors three policy stopping criteria.
First, \textbf{Fatal Minefield Interception}, where $\phi^{\text{fatal}}(e_k)=1$, signals immediately on the raw event clock upon hazardous or irreversible actions.
Second, \textbf{Persistent Low Quality}, where $S_j < \theta_{\text{fail}}$, signals when accumulated stage deficits indicate unrecoverable goal drift.
Third, \textbf{Frontier Stagnation} signals when an agent takes $N_{\text{stag}}$ consecutive actions without advancing the ready frontier $\mathcal{F}_t$, detecting pathological stagnation.
A stop emits an auditable report recording the step, dominator anchor, and failure reason.
\section{Experiments}
\label{sec:experiments}

We evaluate DynSTEER on ToolSandbox and SWE-bench Pro using the same set of instruction-tuned agent backbones. The experiments address four central research questions.
\textbf{RQ1}: whether stage-wise dynamic evaluation provides superior error focus and model discriminability over whole-trajectory evaluation.
\textbf{RQ2}: how much dynamic judge levels, dynamic dimension weighting, and execution-time early stopping contribute to discriminability and efficiency.
\textbf{RQ3}: whether milestone graphs compiled from task inputs reliably capture necessary goals while accommodating diverse solution paths.
\textbf{RQ4}: how effectively execution-time early stopping identifies and removes wasteful execution prefixes.

\subsection{Experimental Setup}
\label{sec:exp_setup}
\textbf{Benchmark Environments and Evaluated Models.} We evaluate on two benchmarks. \textbf{ToolSandbox}~\citep{gu2024toolsandbox} is a stateful tool-use benchmark that scores complete trajectories through milestone and minefield checks.
\textbf{SWE-bench Pro}~\citep{deng2025swebenchpro} is a repository-level software-engineering benchmark that requires multi-file edits and test-based resolution. Appendix~\ref{app:dataset_stats} reports the dataset statistics, and Appendix~\ref{app:experiment_config} details task filtering, baseline implementation, runtime parameters, and ablation variants.
We evaluate \textbf{DeepSeek-V4-Pro}, \textbf{DeepSeek-V4-Flash}~\citep{deepseekai2024deepseekv3}\footnote{The experiments use the post-2026-07-31 version of DeepSeek-V4-Flash.}, \textbf{Qwen3-Max-2026-01-23}, and \textbf{Qwen-Plus-2025-12-01}~\citep{yang2024qwen25} under standard ReAct prompting~\citep{yao2023react}. We set the evaluator model uniformly to Qwen3-Max-2026-01 with a temperature of 0.2.

\textbf{Comparative Paradigms.} For controlled comparison, \textbf{\textsc{Whole-Traj}} and \textbf{\textsc{DynSTEER}} are evaluated on the same source trajectory. \textsc{Whole-Traj} treats the complete trajectory as one evaluation unit, whereas DynSTEER applies dynamic stage-wise evaluation to the same recorded trajectory. The details are specified in Appendix~\ref{app:experiment_config}.

\textbf{Evaluation Metrics.} We assess performance across three dimensions:
\begin{itemize}[leftmargin=*,itemsep=1.5pt,topsep=1pt]
    \item \textbf{Metric Discriminability:} Following \citet{qian2026benchmark}, we measure Discriminability Score (\textbf{DS}). 
    DS quantifies the ability to separate models by combining score dispersion with the fraction of model pairs whose mean scores differ by more than a margin:
    \begin{equation}
        \text{DS}(\epsilon) = \left( \frac{\sigma_{\text{pop}}}{\mu} \right) \cdot \sqrt{\frac{N_{\text{sig}}(\epsilon)}{\binom{M}{2}}}, \quad N_{\text{sig}}(\epsilon) = \sum_{1 \le i < j \le M} \mathbb{I}\left( |\bar{s}_{m_i} - \bar{s}_{m_j}| > \epsilon \right)
        \label{eq:ds}
    \end{equation}
    where $\mu$ and $\sigma_{\text{pop}}$ are the mean score and standard deviation across models, and $N_{\text{sig}}(\epsilon)$ is the number of statistically distinguishable model pairs at margin $\epsilon$. For readability, we normalize agent scores to $[0, 100]$ and report all DS results with $\epsilon = 1.00$.
    \item \textbf{Milestone Graph Quality:} On ToolSandbox, we compute Tool Operation Micro-F1 and Macro-F1 to measure semantic tool coverage, Fatal Minefield F1 for safety-critical action detection, and Graph Edit Distance (GED)-based similarity~\citep{sanfeliu1983distance} for structural topology alignment.
    \item \textbf{Efficiency and Early Termination:} For each model, \textbf{Saved Steps\%} is the percentage of matched Whole-Traj full-trajectory steps avoided by DynSTEER policy stops. \textbf{Stop Rate} divides policy stops by the matched evaluation population. For DynSTEER, \textbf{Judge LLM Calls\%} is the ratio of stage-evaluation LLM calls to agent execution steps.
\end{itemize}

\subsection{RQ1: Stage-wise Dynamic Evaluation, Error Focus, and Model Discriminability}
\label{sec:exp_rq1}

\begin{table*}[t]
\centering
\small
\caption{Main Evaluation Results on ToolSandbox and SWE-bench Pro.}
\label{tab:main_results}
\vspace{0.15cm}
\resizebox{\textwidth}{!}{%
\begin{tabular}{llcccccc}
\toprule
\textbf{Benchmark} & \textbf{Evaluator} & \textbf{Model} & \textbf{Mean Score} $\uparrow$ & \textbf{Stop Rate} & \textbf{DS ($\epsilon{=}1.00$)} $\uparrow$ & \textbf{Saved Steps\%} $\uparrow$ & \textbf{Judge LLM Calls\%} \\
\midrule
\multirow{8}{*}{SWE-bench Pro}
& \multirow{4}{*}{\textsc{Whole-Traj.}}
 & DeepSeek-V4-Flash & $69.54 \pm 8.60$ & 0.00\% & \multirow{4}{*}{0.0324} & 0.00\% & -- \\
 & & DeepSeek-V4-Pro & $63.64 \pm 8.93$ & 0.00\% & & 0.00\% & -- \\
 & & Qwen3-Max-2026-01 & $66.10 \pm 4.15$ & 0.00\% & & 0.00\% & -- \\
 & & Qwen-Plus-2025-12 & $64.03 \pm 3.96$ & 0.00\% & & 0.00\% & -- \\
\cmidrule(lr){2-8}
& \multirow{4}{*}{\textsc{DynSTEER}}
 & DeepSeek-V4-Flash & $77.40 \pm 5.50$ & 6.12\% & \multirow{4}{*}{\textbf{0.0857}} & 12.19\% & $9.60 \pm 1.12$\% \\
 & & DeepSeek-V4-Pro & $69.90 \pm 0.62$ & 9.00\% & & 10.78\% & $6.80 \pm 0.37$\% \\
 & & Qwen3-Max-2026-01 & $64.93 \pm 3.98$ & 11.00\% & & 6.64\% & $9.64 \pm 0.47$\% \\
 & & Qwen-Plus-2025-12 & $61.87 \pm 2.87$ & 7.07\% & & 2.38\% & $10.14 \pm 1.06$\% \\
\midrule
\multirow{8}{*}{ToolSandbox}
& \multirow{4}{*}{\textsc{Whole-Traj.}}
 & DeepSeek-V4-Flash & $78.80 \pm 0.43$ & 0.00\% & \multirow{4}{*}{0.0270} & 0.00\% & -- \\
 & & DeepSeek-V4-Pro & $77.18 \pm 0.14$ & 0.00\% & & 0.00\% & -- \\
 & & Qwen3-Max-2026-01 & $73.68 \pm 1.00$ & 0.00\% & & 0.00\% & -- \\
 & & Qwen-Plus-2025-12 & $73.62 \pm 1.09$ & 0.00\% & & 0.00\% & -- \\
\cmidrule(lr){2-8}
& \multirow{4}{*}{\textsc{DynSTEER}}
 & DeepSeek-V4-Flash & $73.38 \pm 0.26$ & 16.63\% & \multirow{4}{*}{\textbf{0.0500}} & 11.44\% & $17.97 \pm 0.66$\% \\
 & & DeepSeek-V4-Pro & $71.11 \pm 0.21$ & 17.68\% & & 16.06\% & $20.69 \pm 0.41$\% \\
 & & Qwen3-Max-2026-01 & $65.03 \pm 0.94$ & 27.24\% & & 21.54\% & $20.03 \pm 0.20$\% \\
 & & Qwen-Plus-2025-12 & $66.15 \pm 0.96$ & 18.53\% & & 22.14\% & $21.33 \pm 0.18$\% \\
\bottomrule
\end{tabular}%
}
\end{table*}

\textbf{Discrimination and efficiency.} DynSTEER substantially enlarges the measurable capability margin: DS increases from 0.0270 to \textbf{0.0500} on ToolSandbox and from 0.0324 to \textbf{0.0857} on SWE-bench Pro, yielding relative gains of 85.2\% and 164.0\%.
This change separates all six model pairs and raises the standard deviation of model means from 2.24 to 3.45 and from 2.34 to 5.87, respectively.

The pattern is consistent with stage-wise review: DynSTEER traces frontier progress and assigns lower stage-level evidence where an agent repeatedly fails tool-state transitions, violates safety constraints, or stalls without progress.
Consequently, weak process behavior is no longer hidden behind an ultimately resolved outcome, and stronger models recover visibly from weaker ones in the score distribution.
Because comparisons are paired on identical source trajectories, these gains reflect evaluator sensitivity to process quality rather than differences in the evaluated rollouts.

DynSTEER also removes 17.74\% and 8.40\% of Whole-Traj execution steps, achieving sharper discrimination while reducing evaluation execution cost. 
Additionally, compared with atomic-step evaluation, DynSTEER's judge LLM calls account for roughly 10--20\% of total agent execution steps, indicating that it substantially reduces the cost of invoking LLM in evaluation.

\textbf{Saved Steps vary substantially across backbones.}
Across benchmarks, model-level savings range from 11.44--22.14\% on ToolSandbox to 2.38--12.19\% on SWE-bench Pro.
The narrower and lower SWE-bench Pro range reflects fewer policy stops in the smaller repository-level population, although individual stopped prefixes can still consume substantial execution budget.
These results demonstrate that DynSTEER removes prefixes judged unrecoverable without collapsing model separation.

\textbf{Stability and rank consistency.} The mean repeat-level standard deviation decreases from 0.66 to 0.59 on ToolSandbox and from 6.41 to 3.24 on SWE-bench Pro.
Within each benchmark, DynSTEER preserves five of six pairwise model orders on ToolSandbox (Kendall $\tau=0.67$) and four of six on SWE-bench Pro ($\tau=0.33$); DeepSeek-V4-Flash remains first in both cases.
In summary, DynSTEER offers greater model discriminability and more stable results across repeated evaluations.

\subsection{RQ2: Component Contributions to Discriminability and Early Stopping}
\label{sec:exp_rq2}

\begin{table*}[t]
\centering
\small
\caption{Component Ablation on a ToolSandbox Subset.}
\label{tab:ablation_results}
\vspace{0.15cm}
\resizebox{0.7\textwidth}{!}{%
\begin{tabular}{lccc}
\toprule
\textbf{Configuration} & \textbf{DS ($\epsilon{=}1.00$)} $\uparrow$ & \textbf{Stop Rate} & \textbf{Saved Steps\%} \\
\midrule
\textsc{Whole-Traj.} & 0.0228 & 0.00\% & --- \\
\textsc{DynSTEER} & \textbf{0.0528} & \textbf{18.92\%} & \textbf{12.90\%} \\
\quad\textit{-w/o} dynamic weighting & 0.0414 & 17.92\% & 12.75\% \\
\quad\textit{-w/o} dynamic judge level & 0.0439 & 15.83\% & 12.58\% \\
\quad\textit{-w/o} dynamic & 0.0311 & 15.83\% & 12.49\% \\
\quad\textit{-w/o} policy stop & 0.0286 & 0.00\% & --- \\
\bottomrule
\end{tabular}%
}
\end{table*}

To answer RQ2, we conduct a component ablation on a ToolSandbox subset, and the sampling protocol and ablated variants are detailed in Appendix~\ref{app:experiment_config}.
As shown in Table~\ref{tab:ablation_results}, removing any dynamic component or the early-stopping mechanism hurts the performance.
Nevertheless, every ablated DynSTEER configuration remains more discriminative than \textsc{Whole-Traj}.
This uniform improvement over whole-trajectory scoring indicates that stage-wise organization itself helps separate models and expose deficiency patterns, while the dynamic components further enlarge that margin.

\textbf{Early stopping has the largest single-component impact.} Disabling policy stopping reduces DS from 0.0528 to 0.0286. A plausible reason is that early stopping converts an unrecoverable execution prefix into decisive failure evidence.
This binary signal complements graded stage scores: without it, a long routine suffix can dilute the score distribution and make systematically weak executions resemble recoverable noise.

\textbf{Dynamic weighting and judge levels contribute complementary separation signals.} Dynamically adjusting dimension weights makes models easier to distinguish: it reallocates evidence toward low-scoring or high-uncertainty dimensions and prevents uniform averaging from canceling complementary weaknesses.
Dynamic judge levels nevertheless provide strong discriminability: relative to stage-only evaluation (\textit{-w/o} dynamic), retaining judge-level escalation under \textit{-w/o} dynamic weighting raises DS from 0.0311 to 0.0414.
The early-stop statistics further separate the two mechanisms: \textit{-w/o} dynamic weighting attains a higher Stop Rate and Saved Steps than \textit{-w/o} dynamic judge level.
This suggests that stronger review of poorly executed dimensions is more likely to expose problems and assign low scores promptly, allowing the agent to stop before additional wasted steps accumulate.

\subsection{RQ3: Milestone Graph Generation Reliability}
\label{sec:exp_rq3}

\begin{table}[t]
\centering
\small
\caption{Evaluation of simulated milestone graphs compiled by DynSTEER against the expert reference graphs on ToolSandbox.}
\label{tab:milestone_gen}
\vspace{0.15cm}
\begin{tabular}{lccc}
\toprule
\textbf{Evaluation Dimension} & \textbf{Precision} & \textbf{Recall} & \textbf{F1 Score} \\
\midrule
Tool Operation (Micro) & 84.0\% & 81.9\% & 82.9\% \\
Tool Operation (Macro) & 90.2\% & 88.4\% & 87.3\% \\
Fatal Minefield Detection & 91.8\% & 75.3\% & 82.7\% \\
\midrule
\textbf{Structural Topology Metric} & \multicolumn{3}{c}{\textbf{Score}} \\
\midrule
Topological GED Similarity & \multicolumn{3}{c}{0.789} \\
Valid DAG Compilation Rate & \multicolumn{3}{c}{100.0\%} \\
\bottomrule
\end{tabular}
\end{table}

Table~\ref{tab:milestone_gen} summarizes the semantic accuracy, structural distance, and safety alignment of generated milestone graphs against canonical expert annotations.
DynSTEER's multi-candidate synthesis and majority consensus achieve an \textbf{82.9\% Tool Operation Micro-F1} and an \textbf{87.3\% Tool Operation Macro-F1}, extending reliable tool coverage beyond frequent operations to tail tools.
Fatal Minefield Detection reaches an \textbf{82.7\% F1}, preferentially preserving irreversible operations from available task inputs.
Transitive reduction retains alternative parallel branches without penalizing valid exploratory variations, such as searching by contact name rather than date, achieving a 100\% valid DAG compilation rate with stronger topological alignment.

These expert-referenced results support using the compiled milestone graph as a principled basis for stage partitioning. The Macro-F1 (87.3\%) and GED similarity (0.789) show that the compiled graphs capture tail tools and fine-grained topology.
Nevertheless, the perfect valid-DAG rate and branch-preserving transitive reduction indicate that compilation yields stable checkpoints without collapsing alternative valid plans across diverse execution strategies.

\subsection{RQ4: Early-Stop Behavior and Counterfactual Savings}
\label{sec:exp_rq4}

\begin{figure*}[t]
\centering
\resizebox{\textwidth}{!}{%
\begin{tikzpicture}[
    font=\small,
    value/.style={font=\scriptsize, inner sep=1.5pt},
    benchmark/.style={font=\scriptsize\bfseries},
    panel/.style={font=\small\bfseries},
    tick/.style={font=\scriptsize, black},
    axis/.style={draw=black!55, line width=0.5pt},
    gridline/.style={draw=black!10, line width=0.4pt},
]
    \definecolor{ScoreBlue}{RGB}{84,124,181}
    \definecolor{StagnationOrange}{RGB}{235,153,73}
    \definecolor{MinefieldRed}{RGB}{193,78,82}

    \fill[ScoreBlue] (0.05,4.55) rectangle ++(0.20,-0.16);
    \node[anchor=west] at (0.37,4.47) {Persistent Low Quality (\textsc{Score})};
    \fill[StagnationOrange] (5.35,4.55) rectangle ++(0.20,-0.16);
    \node[anchor=west] at (5.67,4.47) {Frontier Stagnation};
    \fill[MinefieldRed] (9.15,4.55) rectangle ++(0.20,-0.16);
    \node[anchor=west] at (9.47,4.47) {Fatal Minefield Interception (\textsc{Minefield})};

    \node[panel] at (2.05,3.82) {(a) Stopped Runs (\%)};

    \fill[ScoreBlue] (0.90,1.78) -- ++(90:0.88)
        arc[start angle=90, end angle=60, radius=0.88] -- cycle;
    \fill[StagnationOrange] (0.90,1.78) -- ++(60:0.88)
        arc[start angle=60, end angle=-180, radius=0.88] -- cycle;
    \fill[MinefieldRed] (0.90,1.78) -- ++(-180:0.88)
        arc[start angle=-180, end angle=-270, radius=0.88] -- cycle;
    \node[value, anchor=west] at (1.19,2.84) {8.3\%};
    \node[value] at (1.22,1.23) {66.7\%};
    \node[value, text=white] at (0.48,2.23) {25.0\%};
    \node[benchmark] at (0.90,0.46) {SWE-bench Pro};

    \fill[ScoreBlue] (3.18,1.78) -- ++(90:0.88)
        arc[start angle=90, end angle=34.835, radius=0.88] -- cycle;
    \fill[StagnationOrange] (3.18,1.78) -- ++(34.835:0.88)
        arc[start angle=34.835, end angle=-49.127, radius=0.88] -- cycle;
    \fill[MinefieldRed] (3.18,1.78) -- ++(-49.127:0.88)
        arc[start angle=-49.127, end angle=-270, radius=0.88] -- cycle;
    \node[value, text=white] at (3.59,2.35) {15.3\%};
    \node[value] at (3.99,1.66) {23.3\%};
    \node[value, text=white] at (2.75,1.58) {57.6\%};
    \node[benchmark] at (3.18,0.46) {ToolSandbox};

    \node[panel] at (7.15,3.82) {(b) Unique Tasks (\%)};
    \draw[axis] (5.20,0) -- (9.05,0);
    \draw[axis] (5.20,0) -- (5.20,2.75);
    \foreach \y/\label in {0/0, 1.0/10, 2.0/20} {
        \draw[axis] (5.14,\y) -- (5.20,\y);
        \node[tick, anchor=east] at (5.07,\y) {\label};
    }
    \draw[gridline] (5.20,1.0) -- (9.05,1.0);
    \draw[gridline] (5.20,2.0) -- (9.05,2.0);

    \fill[ScoreBlue] (5.42,0) rectangle (5.64,0.513);
    \fill[StagnationOrange] (5.90,0) rectangle (6.12,1.282);
    \fill[MinefieldRed] (6.38,0) rectangle (6.60,0.769);
    \node[value, anchor=south] at (5.53,0.55) {5.1};
    \node[value, anchor=south] at (6.01,1.32) {12.8};
    \node[value, anchor=south] at (6.49,0.81) {7.7};
    \node[benchmark] at (6.01,-0.31) {SWE-bench Pro};

    \fill[ScoreBlue] (7.52,0) rectangle (7.74,1.257);
    \fill[StagnationOrange] (8.00,0) rectangle (8.22,2.456);
    \fill[MinefieldRed] (8.48,0) rectangle (8.70,1.906);
    \node[value, anchor=south] at (7.63,1.30) {12.6};
    \node[value, anchor=south] at (8.11,2.50) {24.6};
    \node[value, anchor=south] at (8.59,1.95) {19.1};
    \node[benchmark] at (8.11,-0.31) {ToolSandbox};

    \node[panel] at (12.80,3.82) {(c) Saved Steps (\%)};
    \draw[axis] (10.75,0.25) -- (10.75,3.20);

    \fill[ScoreBlue] (10.75,2.33) rectangle (11.08,2.55);
    \fill[StagnationOrange] (10.75,2.03) rectangle (12.35,2.25);
    \fill[MinefieldRed] (10.75,1.73) rectangle (11.59,1.95);
    \node[value, anchor=west] at (11.14,2.44) {1.0};
    \node[value, anchor=west] at (12.41,2.14) {4.9};
    \node[value, anchor=west] at (11.65,1.84) {2.5};
    \node[benchmark, anchor=east, align=right] at (10.59,2.14) {SWE-bench\\ Pro};

    \fill[ScoreBlue] (10.75,1.03) rectangle (11.04,1.25);
    \fill[StagnationOrange] (10.75,0.73) rectangle (12.24,0.95);
    \fill[MinefieldRed] (10.75,0.43) rectangle (14.82,0.65);
    \node[value, anchor=west] at (11.10,1.14) {0.9};
    \node[value, anchor=west] at (12.30,0.84) {4.5};
    \node[value, anchor=west] at (14.88,0.54) {12.3};
    \node[benchmark, anchor=east] at (10.59,0.84) {ToolSandbox};
\end{tikzpicture}%
}
\caption{\textbf{Early-stop behavior and savings by termination criterion.} (a) Stopped-run shares, (b) unique-task coverage, and (c) Saved Steps contributions. Unique-task shares are not additive.}
\label{fig:early_stop}
\end{figure*}
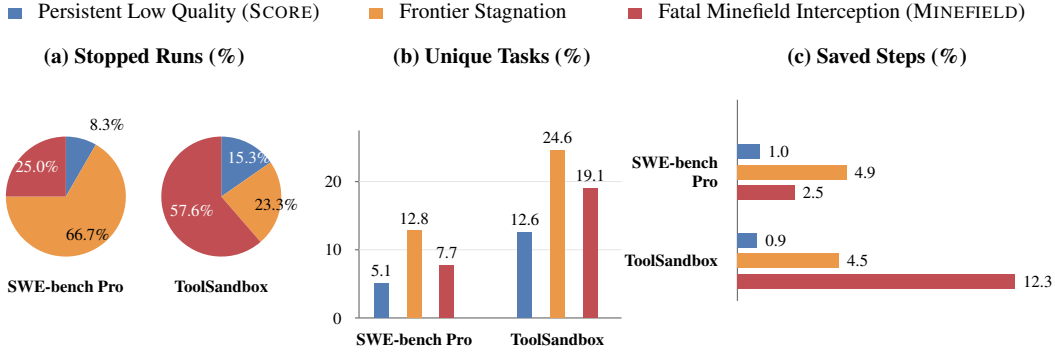

\textbf{Failure modes analysis.} Figure~\ref{fig:early_stop}(a) shows that Fatal Minefield Interception dominates ToolSandbox stopped runs at 57.56\%, and Figure~\ref{fig:early_stop}(c) shows its 12.34\% contribution to the aggregate 17.74\% Saved Steps.
Its unique-task share is only 19.06\%, whereas Frontier Stagnation covers the largest unique-task share (24.56\%) but saves only 4.53\%.
Thus, ToolSandbox failures often occur early as unsafe or invalid calls on interactive tasks.

On SWE-bench Pro, Frontier Stagnation dominates stopped runs at 66.67\% and unique-task coverage at 12.82\%, whereas Fatal Minefield Interception accounts for 25.00\% and 7.69\%, respectively (Figure~\ref{fig:early_stop}(a,b)). Frontier Stagnation contributes the largest saving, 4.86\% of the aggregate 8.40\%, reflecting prolonged repository exploration and repair loops before a terminal test result. Persistent Low Quality remains conservative, saving 1.00\% on SWE-bench Pro and 0.87\% on ToolSandbox.

\textbf{Common task-level stagnation.} Despite these distributional differences, Frontier Stagnation has the largest unique-task coverage on both benchmarks: 12.82\% on SWE-bench Pro and 24.56\% on ToolSandbox (Figure~\ref{fig:early_stop}(b)). Its stopped-run share and contribution to savings still differ across datasets, but the pattern is consistent: stagnation arises in more distinct tasks than in either Persistent Low Quality or Fatal Minefield Interception. This indicates that the most prevalent risk is not confined to immediate safety violations or uniformly poor execution. Instead, agents on both benchmarks frequently enter extended progress vacuums in which local actions continue without advancing the task frontier, making frontier-aware execution-time review valuable in both tool-use and repository-level settings.
In addition, Appendix~\ref{app:case_studies} presents cases illustrating auditable stop decisions.

\section{Conclusion}
\label{sec:conclusion}

In this work, we propose \textbf{DynSTEER}, a dynamic stage-wise framework that advances LLM agent trajectory evaluation from passive, post-hoc terminal verification toward proactive execution review.
DynSTEER segments rollouts at key execution nodes, adapts its multi-level review strategy to stage-level results, compiles a path-tolerant milestone graph without reference leakage, and terminates unrecoverable executions.
Experimental results show that DynSTEER improves evaluation discriminability by over 85\% compared with whole-trajectory evaluation and saves 17.74\% of execution steps, providing a principled and cost-aware foundation for LLM-agent evaluation and optimization.


\bibliography{references}
\bibliographystyle{iclr2027_conference}

\newpage
\appendix
\section{Benchmark Dataset Details}
\label{app:dataset_stats}

\textbf{ToolSandbox.} ToolSandbox~\citep{gu2024toolsandbox} is an interactive, stateful benchmark engineered to evaluate complex multi-turn decision-making in tool-using LLM agents. Unlike static benchmarks that rely solely on mock API dictionaries, ToolSandbox maintains a stateful world environment with persistent entities, temporal relationships, and strict data invariants. Table~\ref{tab:dataset_stats} summarizes the evaluated scenario distribution.

\begin{table*}[t]
\centering
\small
\caption{\textbf{ToolSandbox Benchmark Dataset Statistics.} Distribution of evaluated scenarios across functional domains, interaction modalities, adversarial perturbation variants, and safety minefield configurations.}
\label{tab:dataset_stats}
\vspace{0.15cm}
\resizebox{\textwidth}{!}{%
\begin{tabular}{lcccc}
\toprule
\textbf{Scenario Domain / Sub-category} & \textbf{Count} & \textbf{Avg. User Turns} & \textbf{Available Tools} & \textbf{Minefield Invariants} \\
\midrule
Calendar \& Scheduling Management & 112 & 3.4 & 8 -- 18 & Data integrity, date bounds \\
Messaging \& Notification Services & 136 & 2.9 & 6 -- 15 & Privacy, unverified sends \\
Contact \& Personal CRM & 98 & 2.6 & 5 -- 12 & Irreversible deletion \\
Financial \& Transaction Processing & 84 & 4.1 & 10 -- 24 & Unauthorized transfer \\
Information Retrieval \& Weather/Geocoding & 79 & 3.2 & 7 -- 16 & Parameter type constraints \\
\midrule
\textbf{Perturbation \& Difficulty Variants} & \textbf{Count} & \multicolumn{3}{c}{\textbf{Properties \& Challenge Characteristics}} \\
\midrule
Standard Canonical & 127 & \multicolumn{3}{l}{Nominal parameters, standard system prompt and API schemas} \\
Distraction Tools (3--10 extra APIs) & 158 & \multicolumn{3}{l}{Noise tools added to prompt to test agent tool retrieval} \\
Scrambled Arguments / Descriptions & 142 & \multicolumn{3}{l}{Shuffled arg orders and adversarial field descriptors} \\
Insufficient Information (Rejection Tasks) & 82 & \multicolumn{3}{l}{Ambiguous queries requiring clarification or safe refusal} \\
\midrule
\textbf{Total Evaluated Benchmark Pool} & \textbf{509} & \textbf{3.2 turns} & \textbf{12.4 tools/task} & \textbf{704 total minefields} \\
\bottomrule
\end{tabular}%
}
\end{table*}

\textbf{SWE-bench Pro.} SWE-bench Pro~\citep{deng2025swebenchpro} supplies repository-level software-engineering tasks with Docker execution environments, gold patches, and test-based resolution. Its public test split contains 731 tasks from 11 repositories. Table~\ref{tab:swebench_dataset_stats} reports the repository-stratified sample used in our experiments.

\begin{table}[t]
\centering
\small
\caption{\textbf{SWE-bench Pro Public-Split Sampling Statistics.} The evaluated subset is stratified by repository to preserve the public test split's project diversity.}
\label{tab:swebench_dataset_stats}
\vspace{0.15cm}
\begin{tabular}{lcc}
\toprule
\textbf{Repository} & \textbf{Public Split} & \textbf{Stratified Sample} \\
\midrule
NodeBB/NodeBB & 44 & 6 \\
ansible/ansible & 96 & 14 \\
element-hq/element-web & 56 & 7 \\
flipt-io/flipt & 85 & 12 \\
future-architect/vuls & 62 & 8 \\
gravitational/teleport & 76 & 11 \\
internetarchive/openlibrary & 91 & 13 \\
navidrome/navidrome & 57 & 7 \\
protonmail/webclients & 65 & 9 \\
qutebrowser/qutebrowser & 79 & 11 \\
tutao/tutanota & 20 & 2 \\
\midrule
\textbf{Total} & \textbf{731} & \textbf{100} \\
\bottomrule
\end{tabular}
\end{table}

\section{Detailed Experimental Configuration}
\label{app:experiment_config}

This appendix documents task selection, the benchmark-specific \textsc{Whole-Traj} implementation, runtime parameters, and component ablations. All experiments use the four agent backbones in Section~\ref{sec:exp_setup} and three repeats per task. DynSTEER evaluations replay the same complete source trajectories as \textsc{Whole-Traj}; a virtual stop marks a counterfactual prefix boundary and does not launch a separate rollout.

\subsection{Task Selection and Filtering}
\label{app:task_selection}

\textbf{ToolSandbox.} The main experiment uses all 509 ToolSandbox scenarios backed by fully local, sandboxed execution environments rather than live RapidAPI endpoints. This selection prioritizes: (1) reproducibility, because external endpoints introduce downtime, schema drift, network latency, and rate limits; (2) inspectable state snapshots and invariant audits, which are required for stage settlement and safety analysis; and (3) hermetic execution, which avoids external data leakage and uncontrolled API cost.

\textbf{SWE-bench Pro.} We start from the public test split with 731 tasks from 11 repositories and draw a repository-stratified sample of 100 tasks (Table~\ref{tab:swebench_dataset_stats}). Task containers use only local Docker images or case-level cached images. When a required image cannot be loaded, the run is an infrastructure failure rather than evidence of agent capability; we remove the affected \textsc{Whole-Traj}/DynSTEER pair before computing score, stop-rate, DS, and early-stop statistics. This leaves 39 matched tasks. The retained set preserves repository stratification at sampling time but is smaller than the complete public split, which we report as a limitation.

\textbf{Component-ablation subset.} The ablation samples 100 ToolSandbox scenarios from the 509 main-experiment scenarios, stratified by task type and safety category. It uses the same four backbones, three repeats, source trajectories, and paired-comparison protocol as the main experiment.

\subsection{\textsc{Whole-Traj} Implementation}
\label{app:whole_traj_details}

\textsc{Whole-Traj} is the non-stage-wise comparison paradigm: it receives the complete trajectory as one unit and does not partition evidence by dominator milestones.

On ToolSandbox, it calls the benchmark-native trajectory evaluator over the recorded execution context. The evaluator compares milestone and minefield evidence exposed by the trajectory and returns a single similarity score, normalized to the common $[0,100]$ reporting scale.

On SWE-bench Pro, the official benchmark protocol resolves a patch with tests but does not provide a native whole-trajectory evaluator. We therefore use an outcome-informed whole-trajectory audit. A fixed Qwen3-Max-2026-01-23 auditor (temperature $0.2$) receives the task, repository context, complete command/result trajectory, final patch, and terminal patch/test outcome. It returns one holistic score on the common reporting scale that audits execution quality conditioned on the terminal outcome. This construction provides baseline access to both the full trajectory and the benchmark's terminal evidence while preserving single-unit evaluation granularity.

\subsection{Runtime Parameters}
\label{app:runtime_parameters}

ToolSandbox agents use standard ReAct prompting, a maximum of 100 messages, and a fixed Qwen-Plus simulated user. SWE-bench Pro agents use a shell tool, at most one command per turn, at most 60 tool calls, a 600-second command timeout, and model temperature $0.0$. Agent credentials are injected only into clients and are excluded from trajectories and released artifacts.

Milestone graphs are generated with Qwen3-Max-2026-01-23 at temperature $0.0$, targeting 6 candidate graphs with at most 4 generation batches. 
LLM judges use the same fixed judge model at temperature $0.2$. 
Unless otherwise specified, stage thresholds are pass $0.80$, warn $0.60$, fail $0.40$, low dimension uncertainty $0.20$, high dimension uncertainty $0.45$, fatal-minefield threshold $0.95$, and threshold margin $0.05$. 
Dynamic-weight sensitivities are $\alpha=\beta=1.0$, and frontier stagnation fires after 16 observations without ready-frontier progress.

\subsection{Component-Ablation Definitions}
\label{app:ablation_details}

All ablations retain the same source trajectories, milestone graphs, minefield checks, and models.

\begin{itemize}[leftmargin=*,itemsep=1.5pt,topsep=2pt]
    \item \textbf{\textit{-w/o} dynamic weighting:} dimension weights remain uniform, and adaptive judge-level routing and policy stopping remain enabled.
    \item \textbf{\textit{-w/o} dynamic judge level:} adaptive routing is disabled, and every dimension uses the fixed cheap structural judge. Dynamic weighting and policy stopping remain enabled.
    \item \textbf{\textit{-w/o} dynamic:} both adaptive routing and dynamic weighting are disabled. Evaluation uses uniform weights and the fixed cheap judge, while policy stopping remains enabled.
    \item \textbf{\textit{-w/o} policy stop:} review and scoring are unchanged, but all stopping criteria are suppressed. Replay covers the full source trajectory, so Saved Steps is not applicable.
\end{itemize}
\section{Algorithmic Formulations and Proof Details}
\label{app:algorithms}

\subsection{DynSTEER Execution-Time Stage Evaluation and Control Algorithm}
Algorithm~\ref{alg:dynsteer} formalizes execution-time stage evaluation, dominator evidence-window extraction, uncertainty-adaptive judge-level adaptation, and early termination. The pseudocode separates the closed-action clock $t$ from the evaluated-stage clock $j$: unmatched actions can still trigger frontier-stagnation monitoring without advancing the weight or judge-policy index.

\begin{algorithm}[t]
\caption{DynSTEER Execution-Time Stage Evaluation and Control}
\label{alg:dynsteer}
\begin{algorithmic}[1]
\REQUIRE Task inputs $V_{\text{task}}$, rollout events $\langle e_1,\dots,e_K\rangle$, candidate count $C$, stop thresholds $\Theta$
\ENSURE Final evaluation report $\mathcal{R}$, termination status
\STATE $\{G_1, \dots, G_C\} \leftarrow \text{SynthesizeCandidates}(V_{\text{task}}, C)$
\STATE $G \leftarrow \text{ValidateAndAggregate}(\{G_1, \dots, G_C\})$ \COMMENT{Milestone DAG $G = (M, E, \mathcal{K}, \mathcal{M}_{\text{fatal}})$}
\STATE $M_{\text{settled}} \leftarrow \emptyset,\; T_D(s_0) \leftarrow 0,\; t \leftarrow 0,\; j \leftarrow 0,\; n_{\text{stag}} \leftarrow 0$
\STATE $\mathcal{R} \leftarrow \emptyset,\; \mathbf{w}_0 \leftarrow \text{InitWeights}(),\; \mathbf{L}_0 \leftarrow \{J_{\text{cheap}}\}_{d \in \mathcal{D}}$
\FOR{each raw event $e_k$ in the rollout}
    \IF{$\text{MatchMinefield}(e_k, \mathcal{M}_{\text{fatal}})$}
        \RETURN $\text{Halt}(\mathcal{R}, \text{reason}=\text{"Fatal Minefield"}, k)$
    \ENDIF
    \IF{$e_k$ closes agent action $a_t$ with observed state $s_t$}
        \STATE $t \leftarrow t + 1$
        \STATE $\mathcal{F}_t \leftarrow \{m \in M \setminus M_{\text{settled}} \mid \text{Pred}(m) \subseteq M_{\text{settled}}\}$
        \STATE $m^*, \mathcal{A}_t \leftarrow \text{MatchMilestone}(\mathcal{F}_t, a_t, s_t)$ \COMMENT{$\mathcal{A}_t$ stores ready-candidate attempt scores}
        \STATE $S_j, \mathbf{s}_j, \mathbf{u}_j, \mathbf{d}_j \leftarrow \mathrm{None}$
        \IF{$m^* \neq \text{None}$}
            \STATE $\mathcal{W} \leftarrow [T_D(\text{idom}(m^*)), t]$ \COMMENT{Dominator-anchored evidence window}
            \STATE $S_j, \mathbf{s}_j, \mathbf{u}_j, \mathbf{d}_j \leftarrow \text{EvaluateStage}(\mathcal{W}, m^*, \mathbf{w}_j, \mathbf{L}_j)$
            \STATE $\mathcal{R} \leftarrow \mathcal{R} \cup \{(m^*, S_j, \mathbf{d}_j)\}$
            \STATE $\mathbf{w}_{j+1}, \mathbf{L}_{j+1} \leftarrow \text{UpdateEvaluationPolicy}(\mathbf{w}_j, \mathbf{s}_j, \mathbf{u}_j, S_j, \Theta)$
            \STATE $j \leftarrow j + 1$
            \IF{$\text{AcceptStage}(S_j, \mathbf{d}_j, \Theta)$}
                \STATE $M_{\text{settled}} \leftarrow M_{\text{settled}} \cup \{m^*\},\; T_D(m^*) \leftarrow t,\; n_{\text{stag}} \leftarrow 0$
                \STATE $\mathcal{F}_t \leftarrow \text{AdvanceFrontier}(\mathcal{F}_t, m^*)$
            \ELSE
                \STATE $n_{\text{stag}} \leftarrow \text{UpdateStagnationWatch}(\mathcal{F}_t, \mathcal{A}_t, n_{\text{stag}}, \Theta)$
            \ENDIF
        \ELSE
            \STATE $n_{\text{stag}} \leftarrow \text{UpdateStagnationWatch}(\mathcal{F}_t, \mathcal{A}_t, n_{\text{stag}}, \Theta)$
        \ENDIF
        \STATE $z_t \leftarrow \text{CheckEarlyStopCriteria}(\mathcal{R}, \mathcal{F}_t, S_j, \mathbf{d}_j, n_{\text{stag}}, t, \Theta)$
        \IF{$z_t \neq \text{None}$}
            \RETURN $\text{Halt}(\mathcal{R}, \text{reason}=z_t.\text{reason}, t)$
        \ENDIF
    \ENDIF
\ENDFOR
\RETURN $\text{FinalizeReport}(\mathcal{R}, M_{\text{settled}}, \text{status}=\text{"Completed"})$
\end{algorithmic}
\end{algorithm}

The sentinel $S_j=\bot$ marks a closed action that produces no stage score; \textsc{CheckEarlyStopCriteria} then skips the low-quality criterion but can still fire frontier stagnation from $\mathcal{A}_t$ and $n_{\text{stag}}$. Thus every argument passed to the stop checker is defined. The map $T_D$ records dominator settlement times, with the augmented entry $s_0$ anchored at zero. An accepted stage requires a ready-milestone match and successful hard-constraint/pass-threshold checks; only accepted stages advance $M_{\text{settled}}$ and reset stagnation. Failed and unmatched attempts remain in the audit trail through $\mathcal{R}$ or $\mathcal{A}_t$, while \textsc{UpdateStagnationWatch} counts consecutive attempts without meaningful frontier-progress improvement.

\subsection{Candidate Contract Validation}
\label{app:contract_validation}

Before consensus, we deterministically check every candidate; we consult no LLM during this filtering step. \textbf{Schema validation} rejects malformed graphs and unknown operations: each milestone must have a unique identifier, a well-formed success condition, and a canonical tool name present in $\mathcal{A}$; minefield descriptors must likewise reference available destructive operations. It also checks that edge endpoints, dependency declarations, and output-field references exist.

\textbf{Structural validation} verifies that the directed precedence relation is finite, free of self-loops and cycles, and consistent with milestone reachability. Edges that duplicate an existing dependency are collapsed before aggregation, while genuinely distinct precedence constraints are retained.

\textbf{Binding validation} requires every dynamic parameter binding to name a milestone that can execute earlier in the same candidate, an output field declared by that milestone, and a target parameter whose schema accepts the referenced type. Literal arguments are checked against required keys and declared types. Thus, the system removes hallucinated APIs, malformed payloads, cyclic plans, and dangling or type-incompatible bindings before majority consensus.

\subsection{Blueprint Consensus Aggregation Details}
The blueprint aggregation pipeline synthesizes multiple candidate graphs $\{G_1, \dots, G_C\}$ into a canonical Milestone DAG $G = (M, E, \mathcal{K}, \mathcal{M}_{\text{fatal}})$. 

Let each candidate graph be represented by its set of nodes $M_c$, directed precedence edges $E_c$, and tool argument bindings $\mathcal{K}_c$. We define node semantic equivalence $\sim$ using canonical tool names and parameter role bindings:
\begin{equation}
    m_i \sim m_j \iff \text{ToolName}(m_i) = \text{ToolName}(m_j) \land \text{Role}(m_i) = \text{Role}(m_j)
\end{equation}
For each equivalence class $[m]$, its consensus support is:
\begin{equation}
    \text{Support}([m]) = \sum_{c=1}^{C_{\text{valid}}} \mathbb{I}\left( \exists m' \in M_c \text{ s.t. } m' \in [m] \right)
\end{equation}
Milestones with $\text{Support}([m]) \ge \lceil (C_{\text{valid}} + 1) / 2 \rceil$ are admitted into the canonical milestone set $M$. Precedence edges are aggregated analogously, and transitive reduction $\text{TR}(E)$ is computed using Tarjan's linear-time topological ordering algorithm~\citep{tarjan1972depth} to eliminate redundant shortcut edges.

\subsection{Immediate Dominator Stage Anchoring Proof}
Let $G = (M, E)$ be the DAG with augmented entry node $s_0$. A node $d$ dominates node $n$ ($d \;\text{dom}\; n$) if every directed path from $s_0$ to $n$ contains $d$. The \textbf{immediate dominator} $\text{idom}(n)$ is the unique node that strictly dominates $n$ without dominating any other strict dominator of $n$. We compute the dominator tree using the Lengauer-Tarjan algorithm~\citep{lengauer1979fast} in $\mathcal{O}(|E| \alpha(|E|, |M|))$ time.

When milestone $m^*$ is settled at step $t$, its stage evidence window $\mathcal{W}(m^*) = [t_{\text{idom}(m^*)}, t]$ encompasses precisely the intermediate steps executed between its immediate dominator and $m^*$. Because $m^*$ could not have been reached without traversing $\text{idom}(m^*)$, all causal dependencies are preserved while concurrent, independent branches are cleanly partitioned.
\section{Theoretical Design of the Dynamic Weighting Mechanism}
\label{app:weighting_rationale}

The evaluator must repeatedly direct attention toward dimensions whose recent evidence reveals either low quality or high uncertainty, while preserving positivity and comparability across stages. 
A normalized exponential tilt is a natural mechanism for this purpose: it turns each observed signal into a multiplicative potential rather than an ad hoc reassignment, so persistent evidence accumulates rather than gets overwritten. Starting from Eq.~\ref{eq:weight_update}, the common factor $\exp(\alpha)$ cancels after normalization and gives
\begin{equation}
    w_d^{(t+1)}
    \propto
    w_d^{(t)}
    \exp\!\left(-\alpha s_d^{(t)}+\beta u_d^{(t)}\right).
    \label{eq:weight_tilt_appendix}
\end{equation}
Thus, each stage multiplies a dimension's weight by the exponential of the dimension-specific potential $-\alpha s_d^{(t)}+\beta u_d^{(t)}$. Lower scores increase the potential's contribution to future attention, and higher uncertainty does the same.

Iterating this multiplicative rule makes the mechanism's cumulative nature explicit. Because normalization introduces only a dimension-independent constant, the ratio between any two dimensions after $T$ stages is
\begin{equation}
    \frac{w_d^{(T+1)}}{w_{d'}^{(T+1)}}
    =
    \frac{w_d^{(1)}}{w_{d'}^{(1)}}
    \exp\!\left[
        \alpha\sum_{t=1}^{T}\left(s_{d'}^{(t)}-s_d^{(t)}\right)
        +
        \beta\sum_{t=1}^{T}\left(u_d^{(t)}-u_{d'}^{(t)}\right)
    \right].
    \label{eq:weight_ratio_appendix}
\end{equation}
Consequently, a linearly growing cumulative score deficit produces an exponentially growing weight ratio. This is the soft-min form of attention: the mechanism does not merely remember the latest weak dimension, but progressively reallocates review capacity toward persistent weaknesses.

To understand what such repeated tilting does to a model's aggregate score, first isolate the score component by freezing a profile $s=(s_1,\dots,s_D)$. Define the partition function, tilted distribution, and tilted score by
\begin{equation}
    Z(\lambda)=\sum_{d\in\mathcal{D}}e^{-\lambda s_d},
    \qquad
    p_d(\lambda)=\frac{e^{-\lambda s_d}}{Z(\lambda)},
    \qquad
    S(\lambda)=\sum_{d\in\mathcal{D}}p_d(\lambda)s_d .
    \label{eq:tilted_score_appendix}
\end{equation}
The parameter $\lambda$ represents the effective score sensitivity accumulated over repeated updates. Differentiating $\log Z$ gives
\begin{equation}
    \frac{\partial}{\partial\lambda}\log Z(\lambda)
    =
    \frac{\sum_d(-s_d)e^{-\lambda s_d}}{Z(\lambda)}
    =
    -S(\lambda),
\end{equation}
and differentiating once more yields
\begin{equation}
    \frac{dS}{d\lambda}
    =
    -\frac{d^2}{d\lambda^2}\log Z(\lambda)
    =
    -\operatorname{Var}_{p(\lambda)}(s)
    \le 0.
    \label{eq:tilt_derivative_appendix}
\end{equation}
This equation contains both the direction and the rate of the score change. Every model's tilted score decreases as focus sharpens, but the decrease is proportional to the weighted variance of its own dimension profile. A flat profile changes little, whereas a profile with a distinct weakness changes sharply.

The next question is what information this tilt exposes. Decompose model $i$'s profile into an average level and a zero-sum deviation:
\begin{equation}
    s_i=m_i\mathbf{1}+v_i,
    \qquad
    \sum_{d\in\mathcal{D}}v_{i,d}=0.
    \label{eq:profile_decomposition_appendix}
\end{equation}
Uniform weighting retains only the first component:
\begin{equation}
    S_i^{\mathrm{avg}}
    =
    \frac{1}{|\mathcal{D}|}\sum_{d\in\mathcal{D}}(m_i+v_{i,d})
    =
    m_i.
\end{equation}
It therefore cancels $v_i$, even though $v_i$ records where a model succeeds and fails. In the sharp-tilt limit, only the weakest dimension receives weight, so
\begin{equation}
    \lim_{\lambda\to\infty}S_i(\lambda)
    =
    \min_{d\in\mathcal{D}}(m_i+v_{i,d})
    =
    m_i+\min_{d\in\mathcal{D}}v_{i,d}.
    \label{eq:softmin_limit_appendix}
\end{equation}
The second term is the depth of the model's worst weakness relative to its own average. For two models $A$ and $B$, the limiting gap becomes
\begin{equation}
    S_A(\infty)-S_B(\infty)
    =
    \underbrace{(m_A-m_B)}_{\text{average-level difference}}
    +
    \underbrace{\left(\min_d v_{A,d}-\min_d v_{B,d}\right)}_{\text{weakness-depth difference}}.
    \label{eq:model_gap_appendix}
\end{equation}
Hence the mechanism changes the comparison axis: uniform weighting compares average levels, whereas exponential tilting progressively adds the information carried by complementary failure patterns. For example, $s_A=(0.9,0.3)$ and $s_B=(0.5,0.9)$ give a uniform gap of $0.100$, but the limiting gap becomes $0.200$ even though both tilted scores decrease. Both models are pulled downward, while the model with the deeper weakness is pulled down faster.

The same reasoning also clarifies when separation is guaranteed. Let $\mathcal{I}$ denote the model set and define the cross-model dispersion
\begin{equation}
    \operatorname{Dis}(\lambda)
    =
    \sum_{i\in\mathcal{I}}
    \left(S_i(\lambda)-\bar S(\lambda)\right)^2,
    \qquad
    \bar S(\lambda)
    =
    \frac{1}{|\mathcal{I}|}\sum_{i\in\mathcal{I}}S_i(\lambda),
\end{equation}
Using Eq.~\ref{eq:tilt_derivative_appendix},
\begin{equation}
    \frac{d\operatorname{Dis}}{d\lambda}
    =
    2\sum_{i\in\mathcal{I}}
    \left(S_i-\bar S\right)\frac{dS_i}{d\lambda}
    =
    -2\sum_{i\in\mathcal{I}}
    \left(S_i-\bar S\right)\operatorname{Var}_{p_i}(s_i).
    \label{eq:dispersion_derivative_appendix}
\end{equation}
where $p_i$ denotes the tilted distribution induced by Eq.~\ref{eq:tilted_score_appendix} for model $i$'s profile $s_i$.

Dispersion increases when models below the current tilted mean have larger internal variances, that is, when their disadvantages are concentrated in deeper weak links. If instead a high-average model also contains the deepest weakness, this derivative can become negative, and the gap can compress. The mechanism therefore guarantees a principled shift of evidence toward weak links, not unconditional divergence between all model pairs.

The uncertainty term now completes the design. In Eq.~\ref{eq:weight_ratio_appendix}, $\beta u_d^{(t)}$ contributes another dimension-specific potential: uncertain evidence receives attention before it is discarded, allowing the next review level to test an apparent weakness with stronger evidence. Once that evidence is resolved, subsequent updates respond to the refined score and uncertainty estimates. Taken together, the score tilt supplies the comparative objective—moving from average performance to weakness depth—while uncertainty controls the epistemic allocation of review capacity during that transition.
\section{Qualitative Case Studies and Error Analysis}
\label{app:case_studies}

\subsection{Case 1: Early Termination of Infinite Search Frontier Stagnation}
\textbf{Scenario ID:} \\
\null\hspace*{1.5em}\nolinkurl{search_message_with_recency_oldest_multiple_user_turn_alt_3_distraction_tools_arg_type_scrambled}

\textbf{Agent Model:} Qwen3-Max-2026-01 under ReAct prompting.

\textbf{Execution Trace Summary:}
\begin{itemize}[leftmargin=*,itemsep=1pt]
    \item The user asks for the oldest matching text. The agent asks one clarification, learns that the oldest message is required, and then obtains the current timestamp; milestone $m_0$ settles with score $1.0$.
    \item The next milestone, retrieving the target message, stays ready but never improves: its best stage score is $0$ for eight consecutive observations.
    \item DynSTEER therefore emits a frontier-stagnation stop at replay observation 43 and maps that decision to the source execution boundary at batch 22.
    \item \textbf{Replay accounting:} the complete source trajectory records 29 execution batches and 41.04s of execution latency; the matched prefix records 22 batches and 24.68s. Stopping there would save 7 batches and 16.36s of execution latency. These are paired counterfactual savings, not a separately executed live interruption.
\end{itemize}

\subsection{Case 2: First-Error Localization vs. Fluke Success}
\textbf{Scenario ID:} \\
\null\hspace*{1.5em}\nolinkurl{add_reminder_content_and_date_and_time_10_distraction_tools}

\textbf{Agent Model:} DeepSeek-V4-Flash.

\textbf{Execution Trace Summary:}
\begin{itemize}[leftmargin=*,itemsep=1pt]
    \item The agent was tasked with setting a reminder with a specific date and time.
    \item In intermediate steps, the agent called an invalid distraction API \texttt{set\_calendar\_alarm} with malformed argument types, which failed.
    \item The agent then stumbled upon \texttt{add\_reminder} after multiple trial-and-error retries.
    \item \textbf{Whole-Traj Evaluation:} Scored the complete trajectory as $1.0000$ because the required milestones were ultimately resolved, while ignoring the 4 erroneous intermediate API calls and high token overhead.
    \item \textbf{DynSTEER Stage Review:} In the DeepSeek-V4-Flash run, the stage reaching $m_0$ scores $0.897$, whereas the subsequent target-message stage falls to $0.272$ and triggers the low-score policy. DynSTEER therefore preserves the distinction between clean execution and an error-prone, inefficient execution.
\end{itemize}

\section{Prompt Templates for Milestone Generation and Stage Judges}
\label{app:prompts}

This section documents excerpts from the primary English prompt templates used by DynSTEER. The excerpts follow the implementation files closely; for long rule lists, we show only representative head/tail rules, and we denote omitted rules with \ldots{}.

\begin{tcolorbox}[colback=gray!4,colframe=black!75,arc=4pt,boxrule=0.8pt,left=8pt,right=8pt,top=6pt,bottom=6pt]
\textbf{Prompt Template: Standard Judge Unified Multi-Field Prompt} \\[-6pt]
\rule{\textwidth}{0.4pt}
\small
You are a strict task-trajectory evaluator. Evaluate only the supplied task, stage interval, constraint\_checks, and trajectory steps. Do not invent external facts.

\textbf{\\Context fields:}
\begin{itemize}[leftmargin=*,itemsep=1pt,topsep=1pt]
  \item \texttt{stage\_goal} is the current stage success condition; \texttt{task.task\_description} is only task background.
  \item \texttt{rubrics} contains only the dimensions to score in this pass; \texttt{dimension\_scores} must contain only those dimensions.
  \item \texttt{steps} are the primary behavioral evidence. Cite step index, actor, event\_type, tool\_call, tool\_result, or raw sender/recipient.
  \item \texttt{constraint\_checks} are auxiliary structured scorer evidence and must not replace auditing steps.
  \item \texttt{interval} provides the evaluated step range, status, milestone\_score, and milestone evidence.
  \item \texttt{required\_output} defines the exact JSON output shape. Do not output \texttt{stage\_score}; code computes it from \texttt{dimension\_scores}.
\end{itemize}

\textbf{\\Evaluation objective:}
\begin{itemize}[leftmargin=*,itemsep=1pt,topsep=1pt]
  \item Decide whether this stage satisfies \texttt{stage\_goal} within the supplied interval.
  \item Score only the dimensions present in \texttt{rubrics} from 0 to 1.
  \item Ground every score in evidence and distinguish completed, partially completed, merely claimed, and unsupported work.
\end{itemize}

\textbf{\\Evidence rules (excerpt):}
\begin{itemize}[leftmargin=*,itemsep=1pt,topsep=1pt]
  \item Every evidence item must reference a step index, \texttt{interval.evidence} item, or \texttt{constraint\_checks} item.
  \item If a judgment cannot be grounded in Context, explain the evidence gap in diagnosis and lower the relevant dimension score.
  \item For \texttt{state\_snapshot} evidence, \texttt{constraint\_checks} may serve as state evidence; user-facing communication quality must still be audited from steps.
  \item \ldots{}
  \item Do not use external facts or assumptions outside Context.
\end{itemize}

Return exactly one JSON object matching \texttt{required\_output}. Do not wrap it in prose.

\textbf{\\Context:} \texttt{\{context\_json\}}
\end{tcolorbox}

\begin{tcolorbox}[colback=gray!4,colframe=black!75,arc=4pt,boxrule=0.8pt,left=8pt,right=8pt,top=6pt,bottom=6pt]
\textbf{Prompt Template: Template of Expensive Specialist Judge Prompts} \\[-6pt]
\rule{\textwidth}{0.4pt}
\small
You are the specialist LLM-Judge for the \texttt{\{dimension\}} dimension.\\
Score only \texttt{\{dimension\_specific\_focus\}}.

\textbf{\\Focus on:}
\begin{itemize}[leftmargin=*,itemsep=1pt,topsep=1pt]
  \item \texttt{\{Dimension-specific forensic criteria\}}.
\end{itemize}

\textbf{\\Evaluation steps:}
\begin{enumerate}[leftmargin=*,itemsep=1pt,topsep=1pt]
  \item Dimension-specific evidence-audit steps.
  \item \ldots{}
  \item The \texttt{dimension\_scores} field may contain only \texttt{\{dimension\}}.
\end{enumerate}
\text{\\}
Use \texttt{Context.rubrics.\{dimension\}} as the score anchor. Return exactly one JSON object matching \texttt{required\_output}.

\textbf{\\Context:} \texttt{\{context\_json\}}
\end{tcolorbox}

\begin{tcolorbox}[colback=gray!4,colframe=black!75,arc=4pt,boxrule=0.8pt,left=8pt,right=8pt,top=6pt,bottom=6pt]
\textbf{Prompt Template: Milestone Graph Candidate Generation Prompt} \\[-6pt]
\rule{\textwidth}{0.4pt}
\small
\textbf{System Prompt:} \\
You generate candidate milestone goal graphs, not step-by-step trajectories. A node belongs in one candidate graph only when, under that candidate's complete and feasible interpretation, every valid completion must reach that operation or goal. One candidate is a hypothesis about necessity; necessity across candidates is decided later by code aggregation.
\\\\\\
\textbf{User Prompt Input:} \\
This is candidate batch \texttt{\{batch\_index\}}. Focus code: \texttt{\{focus\_code\}}. \\
Batch focus: \texttt{\{focus\_instruction\}}

Return exactly \texttt{\{num\_graphs\}} complete, standalone candidate graphs for the current task.
\\\\
\textbf{Independence and diversity rules:}
\begin{itemize}[leftmargin=*,itemsep=1pt,topsep=1pt]
  \item Each graph must define its own dispositions, nodes, edges, and minefields. Local IDs have graph-local scope only.
  \item Seek genuine diversity between two complete judgments: a different visible tool, direct use of public information, a valid bulk state goal, a different necessary dependency, or a justified executability judgment.
  \item \ldots{}
  \item If no genuine alternative is supported, the two graphs may be equivalent. Completeness and correctness take priority over artificial difference.
\end{itemize}

\textbf{\\Current public task JSON:} \texttt{\{task\}}
\\\\
\textbf{Required semantics (excerpt):}
\begin{itemize}[leftmargin=*,itemsep=1pt,topsep=1pt]
  \item \textbf{Necessity, feasibility, and goals:} Keep only milestones that are necessary under this graph's complete judgment. \ldots{} When information is insufficient, mark only the real terminal tool or critical producer needed by the user request; never guess the target object.
  \item \textbf{Node schemas and provenance:} The only allowed node kinds are \texttt{tool\_call}, \texttt{set\_state}, and \texttt{emit\_message}. A \texttt{tool\_call} node has exactly \texttt{local\_id}, \texttt{turn\_id}, \texttt{kind}, \texttt{evidence\_id}, and \texttt{arguments}. \ldots{} Never hard-code or copy a dynamic UUID, row ID, timestamp, coordinate, token, or tool-produced value as \texttt{public\_literal} unless that exact value is explicitly visible in a permitted current source.
  \item \textbf{Edges:} Add an edge only for a true non-commutable data, prerequisite, or goal dependency. \ldots{} No self-loop, duplicate edge, missing endpoint, or cycle is allowed.
  \item \textbf{Minefields:} A minefield marks a fatal wrong tool call that must not occur for the stated turn. Use only severity \texttt{fatal} and the four listed reason codes. \ldots{}
  \item \textbf{Exact response shape:} Return one JSON object with exactly one top-level key, \texttt{graphs}; \texttt{graphs} must contain exactly \texttt{\{num\_graphs\}} graph objects. Every graph object has exactly \texttt{dispositions}, \texttt{nodes}, \texttt{edges}, and \texttt{minefields}. Return raw valid JSON only.
\end{itemize}

\textbf{\\Final response skeleton:} \\
\texttt{\{"graphs":[\{"dispositions":\{\},"nodes":[],} \\
\texttt{"edges":[],"minefields":[]\}, \{"dispositions":\{\},"nodes":[],} \\
\texttt{"edges":[],"minefields":[]\}, ...]\}}
\end{tcolorbox}
\newpage
\section*{AI Use Statement}
\label{appendix: use_llms}
In this work, we used generative AI tools to polish and refine the writing, check grammar, implement methods, assist with mathematical derivations, and assist with translation.

Additionally, we used generative AI tools to draft figures and parts of the paper, summarize and analyze existing literature, search for information, and suggest possible titles and keywords.

We did not use generative AI tools to generate research ideas or conduct data analysis, and the remaining required disclosure tasks do not apply to this work. 

We reviewed all AI-assisted work and carefully checked and verified all LLM-polished content. 

We take responsibility for the final content of this work, including text, claims, or artifacts produced with the aid of generative AI.

\end{document}